\documentclass[10pt,twocolumn,letterpaper]{article}
\usepackage{iccv}
\usepackage{times}
\usepackage{epsfig}
\usepackage{graphicx}
\usepackage{amsmath}
\usepackage{amssymb}
\usepackage{soul}

\usepackage{comment}
\usepackage{mathtools}

\usepackage{graphicx} 
\usepackage{caption,subcaption}
\usepackage{courier}
\newcommand{\museNo}{01070\-421}
\newcommand{\DFVolNo}{PDC\-2022-133642-I00}

\usepackage[pagebackref=true,breaklinks=true,letterpaper=true,colorlinks,bookmarks=false]{hyperref}

\iccvfinalcopy 

\def\iccvPaperID{24}

\ificcvfinal\fi

\newcommand{\minisection}[1]{\vspace{0.04in} \noindent {\bf #1}\ \ }

\begin{document}

%%%%%%%%% TITLE
\title{Continual Evidential Deep Learning for Out-of-Distribution Detection}

\newcommand\refmark[1]{\textsuperscript#1}

\author{
    Eduardo Aguilar\refmark{1}\refmark{2}\refmark{3}\thanks{Corresponding author.}, Bogdan Raducanu\refmark{2}, Petia Radeva\refmark{2}\refmark{3}, Joost Van de Weijer\refmark{2} \\\
    \refmark{1}Dept. de Ingeniería de Sistemas y Computación, Universidad Católica del Norte, Antofagasta, Chile
    \\\
    \refmark{2}Computer Vision Center, Universitat Autònoma de Barcelona, Barcelona, Spain
    \\\
    \refmark{3}Dept. de Matemàtiques i Informàtica, Universitat de Barcelona, Barcelona, Spain
    \\\ \small{\texttt{eduardo.aguilar@ucn.cl, petia.ivanova@ub.edu, \{bogdan,joost\}@cvc.uab.es}}
}

\maketitle
% Remove page # from the first page of camera-ready.
\ificcvfinal\thispagestyle{empty}\fi

%%%%%%%%% ABSTRACT
\begin{abstract}
   Uncertainty-based deep learning models have attracted a great deal of interest for their ability to provide accurate and reliable predictions. Evidential deep learning stands out achieving remarkable performance in detecting out-of-distribution (OOD) data with a single deterministic neural network. Motivated by this fact, in this paper we propose the integration of an evidential deep learning method into a continual learning framework in order to perform simultaneously incremental object classification and OOD detection. 
   Moreover, we analyze the ability of vacuity and dissonance to differentiate between in-distribution data belonging to old classes and OOD data.
   The proposed method \footnote{Code will be made available at \url{https://github.com/Eaaguilart/cedl}.}, called CEDL, is evaluated on CIFAR-100 considering two settings consisting of 5 and 10 tasks, respectively.  
   From the obtained results, we could appreciate that the proposed method, in addition to provide comparable results in object classification with respect to the baseline, largely outperforms OOD detection compared to several posthoc methods on three evaluation metrics: AUROC, AUPR and FPR95. 

\end{abstract}

\section{Introduction}

Deep learning has reached amazing accuracy in a vast number of applications such as image segmentation \cite{kirillov2023sam}, image retrieval \cite{chowdhury2023sbir}, image generation \cite{rombach2022sd}, 3D image reconstruction \cite{neuralangelo2023cvpr}, etc. sometimes outperforming humans in either strategy games \cite{kwon2020alphago} or medical image diagnostics \cite{zhou2021diagnostic}. However, for trustworthy AI methods, it is crucial that the developed algorithms are able to assess the uncertainty of their estimations. Therefore, the field of uncertainty estimation in neural networks has seen a growing research interest \cite{abdar2021uncertaintysurvey}, including methods based on Bayesian Neural Networks \cite{maddox2019bayesian}, Deep Ensembles \cite{rahaman2021uncertainty} and Single Deterministic Neural Networks \cite{van2020uncertainty}. 

Among the methods to estimate the uncertainty of neural networks, \emph{deep evidential learning}~\cite{sensoy2018evidential} stands out because of its ability to pinpoint the various sources of uncertainty: the method can distinguish between a lack of confidence (as measured by \emph{vacuity}) and conflicting evidence (as measured by \emph{dissonance})~\cite{zhao2019quantifying,hu2021multidimensional}. Within this approach, the network's output is used to set the parameters of a Dirichlet distribution on the class probabilities. These parameters can then be used to assess the uncertainty of the network. The method has been successfully evaluated for the task of OOD detection~\cite{sensoy2018evidential}.
However, the method assumes that all the available data is jointly available and can therefore perform the training under the i.i.d. assumption. This assumption is not realistic for many real-world applications, where data arrives sequentially, presenting a distribution shift over time \cite{de2021continual,masana2022class}. In this context, the capability of a model to detect OOD samples has become a necessity in order to guarantee the robustness and safety of these systems. 

On the other hand, continual learning is concerned with the sequential learning from non-stationary data (the data is provided as a sequence of \emph{tasks}), where the learner has to balance the trade-off between the acquisition of new knowledge while consolidating the previous one, a constraint known as \emph{plasticity-stability dilemma} \cite{mermillod2013dilemma}. The main problem of continual learning is represented by \emph{catastrophic forgetting}~\cite{kirkpatrick2017overcoming}, which refers to a significant drop in performance on previous tasks when learning a new task.
Generally, three main strategies exist to address this highly undesirable effect: regularization methods~\cite{kirkpatrick2017overcoming,li2017learning}, rehearsal methods~\cite{rebuffi2017icarl,lopez2017gradient}, and parameter isolation methods~\cite{serra2018overcoming}. These methods are mainly used in combination with a standard cross-entropy loss for the classification task, and when evaluating, a closed-world protocol is applied (meaning evaluation is restricted to classes seen during training). Evaluating these methods within an open-world assumption has only recently received some attention \cite{aljundi2022continual, kim2022collas, kim2022cvprw}. However, to the best of our knowledge, \emph{continual evidential deep learning} has not yet been explored in depth. 

In this paper, we investigate to what extent an Evidential Deep Learning method embedded in a Continual Learning framework allows to detect OOD data and to differentiate it with respect to in-distribution data from old classes. This way, we could better handle different types of data during the learning process and also improve the robustness of the model against OOD data. Our main contributions are the following: 
\begin{itemize}
    \setlength{\itemsep}{3pt}
    \setlength{\parskip}{0pt}
    \setlength{\parsep}{0pt}
    \item We are the first to integrate evidential deep learning into a continual learning approach to simultaneously perform incremental object classification and OOD detection. To this end, we propose a new loss function that combines cross-entropy loss, knowledge distillation and evidential KL-divergence. 
    \item We analyze the ability of vacuity and dissonance to differentiate in-distribution data belonging to old classes from OOD data. Interestingly, we find that vacuity provides a good measure for OOD data detection, while dissonance struggles to distinguish in-distribution (previous classes) data from unseen classes.
    \item With the proposed CEDL method, we have largely outperformed several state-of-the-art posthoc methods used for OOD detection.   
\end{itemize}

\section{Related work}
We will briefly describe the main related works in uncertainty estimation and continual learning.  

\minisection{Uncertainty Estimation.} 
Uncertainty occurs mainly in two types:
epistemic uncertainty, produced by the lack of training data; and aleatoric uncertainty, related to the random noise produced during the sample collection \cite{kendall2017uncertainties}.
To estimate uncertainty, there are several deep neural methods, designed mainly for classification and regression problems \cite{gawlikowski2021survey}, such as: Bayesian methods, which learn a distribution over the weights \cite{blundell2015weight};  Ensemble methods, which arise from their interpretation of Bayesian methods as an ensemble of thin networks with shared weights \cite{lakshminarayanan2017simple}; Test-time augmentation methods, which perform several forward passes by randomly modifying the input data to perform prediction and estimate uncertainty \cite{wang2019aleatoric}; and singles deterministic methods, such as those based on evidence theory \cite{sensoy2018evidential}. 

Uncertainty-aware deep learning models have been successfully developed in the past in a continual learning framework, but with different objectives with respect to this work. Considering that uncertainty is a natural way to identify what to remember and what to change as we continually learn, various works have proposed approaches \cite{nguyen2018variational, ebrahimi2019uncertainty, loo2020generalized} to approximate Bayesian learning with the goal of mitigating catastrophic forgetting by exploiting uncertainty to regularize the parameters of neural networks. On the other hand, single deterministic methods capable of estimating uncertainty have also been proposed as an effective and efficient alternative to Bayesian methods \cite{wiewel2022dirichlet, holmquist2023evidential}. In \cite{wiewel2022dirichlet}, the performance degradation resulting from the distribution shift that occurs due to the limited number of stored samples from the previous task was analyzed. For this purpose, they propose a method based on the Diritchet Priority Network \cite{malinin2018predictive} to model the distribution shift and minimize the bias introduced during the rehearsal process. Recently, an Evidential deep learning approach was proposed for the class-based incremental semantic segmentation problem \cite{holmquist2023evidential}. Here the authors propose to use the estimated uncertainty as a probability measure for the background class, justifying that this class shifts after each increment (past and future classes are correlated). In our work we are going to adopt and evaluate an Evidential deep learning approach to carry out the OOD detection problem.

\minisection{Continual Learning.} There are several surveys on continual learning, focussing on task-incremental learning~\cite{de2021continual}, class-incremental learning (CIL)~\cite{masana2022class}, online continual learning~\cite{mai2022online}, and categorizing underlying biological motivation~\cite{parisi2019continual}. We will here shortly focus on the main strategies developed in continual learning that we apply within the context of this paper. 

Regularization is one of the main strategies to prevent forgetting. Weight regularization methods compute a prior importance for all parameters in the network~\cite{kirkpatrick2017overcoming,liu2018rotate,zenke2017continual}. This prior is then applied to regularize the weights when training new tasks. On the other hand, functional regularization methods~\cite{li2017learning,hou2019learning} apply a regularization on the network output. This regularization is in general based on knowledge distillation~\cite{hinton2014distilling}. 
Another popular strategy is data rehearsal, where a small set of data from previous tasks is stored in a buffer, and replayed during the training of new tasks~\cite{lopez2017gradient, rebuffi2017icarl}. In general, this method is very efficient to counter forgetting, however, it does increase the memory requirements of the system. For settings, where due to privacy concerns or legislation, the storing of data is prohibited, pseudo-rehearsal methods~\cite{shin2017continual,wu2018memory} have been developed which use a generator to replay data from previous tasks. 

In this paper, we propose a method for continual evidential learning that incorporates both regularization and rehearsal to counter catastrophic forgetting.

\minisection{Continual Out-of-Distribution Detection.} OOD has been extensively studied for some time in the classical machine learning setting (when all the available training data is provided in advance) \cite{hendrycks2016baseline,liang2018enhancing,tack2020OOD}. A comprehensive survey on OOD detection could be found in \cite{yang2022surveyOOD} and recently a generalized benchmark has been proposed \cite{yang2022benchmarkOOD} in order to unify the evaluation methodologies of several existing methods. 

However, the study of OOD in the more realistic context of continual learning only recently started to receive attention from the research community, where most of the work adopted the CIL setting. One of the first works who pioneered this research direction is presented in \cite{aljundi2022continual}, where they established benchmarks for the continual OOD detection problem. Additionally, their analysis provided new insights on how potentially forgotten samples are treated by the OOD detection methods. 

A novel approach is proposed in \cite{kim2022collas} where they use exemplars to learn a new and independent classifier that is able to detect OOD samples for each task. In \cite{he2022out} they detect OOD samples by correcting output bias towards new classes and enhancing output confidence difference based on task discriminativeness. In \cite{kim2022cvprw} an unified approach for both class-incremental and task-incremental settings is proposed, using a dual mechanism: (i) task masking to overcome catastrophic forgetting, and (ii) a learning method for building a model for each task based on OOD detection. A different approach is adopted in \cite{rios2022CLOOD}, where they introduce a self-supervised continual novelty detector, which builds incrementally a statistical model over the space of intermediate features produced by a deep network, and utilizes feature reconstruction errors as uncertainty scores to guide the detection of novel samples. 

In \cite{kim2022neurips}, the authors presented a theoretical study according to which an improved CIL performance is strongly correlated with a good OOD detector. This theoretical study has been further extended in \cite{kim2023icml}, where they show that CIL is indeed learnable.

\section{Methodology}

In this section we briefly introduce the evidential deep learning method and then explain our integration of it into a continual learning approach.

\begin{figure*} [!t]
    \centering
    \includegraphics[width=0.85\textwidth]{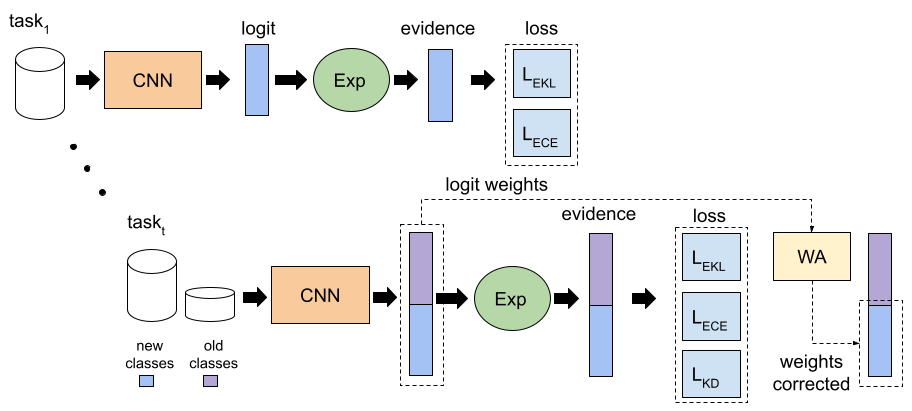}
    \caption{Overview of the proposed Continual Evidential Deep Learning approach.}
    \label{fig:pipeline}
\end{figure*}

\subsection{Evidential Deep Learning}

Deep Learning methods for object recognition often use softmax activation in the output layers to perform classification and provide a measure of confidence. However, it has been demonstrated that the softmax outputs are biased to the data used for training, providing a high probability even for wrong predictions \cite{nguyen2015deep}. To provide a reliable prediction, it is desirable to apply Deep Learning methods that are able to quantify the uncertainty in the predictions. \emph{Evidential Deep Learning} (EDL)  \cite{sensoy2018evidential} is a method developed for this purpose and is able to provide prediction uncertainty using a single deterministic neural network. 
EDL quantifies belief masses and uncertainty using subjective logic theory, which formalizes belief assignments over a discernment framework as a Dirichlet distribution. 

In the evidential framework, it is assumed that the class probability for a sample $i$ is drawn from a prior Dirichlet distribution, given by:
\begin{equation}
    \text{Dir}(p, \alpha) = \frac{\Gamma(S)}{\prod_{c=1}^{C}{\Gamma (\alpha_c)}}\prod_{c=1}^{C}{p_i^{\alpha_c -1}} ; \alpha_c > 0,
\end{equation}
where $p$ is a probability mass function, $C$ is the number of classes, $\alpha = \{\alpha_1, \ldots, \alpha_c\}$ are the Dirichlet parameters, $\Gamma(.)$ is the gamma function and $S = \sum_{c=1}^{C}{\alpha_c}$ is referred to as the Dirichlet strength. As for $\alpha_c$, it is calculated as the sum of the evidence of the c-th class ($e_c$) with one. The evidence is learned by the model and obtained from the logit layer after applying on it a non-negative function such as ReLU \cite{sensoy2018evidential} or Exponential \cite{bao2021evidential}. 

The training of an EDL model for classification is very similar to that of a classical neural network. The main difference is the activation used in the output layer, where softmax is replaced by a non-negative activation. Then, the model can be trained by minimizing the evidential cross-entropy loss ($L_{ECE}$) to form the multi-nomial opinions (learn the evidence) for C-class classification of a given sample $i$ as a Dirichlet distribution. The $L_{ECE}$ is formally defined as follows \cite{sensoy2018evidential, bao2021evidential}:

\begin{equation}
    L_{ECE} = \sum_{j=1}^{C}{y_{ij}(log(S_i) - log(\alpha_{ij}))
    }
\end{equation}
where $y_{i}$ is the one-hot vector encoding of the ground-truth label of the $i$-th sample. 

Note that when the neural network finds shared patterns between different classes, it may generate some evidence about incorrect labels and thus minimize the loss function, $L_{ECE}$. To avoid this behavior, the authors in \cite{sensoy2018evidential} propose a loss function, $L_{EKL}$ which acts as a regularization with the objective that the total evidence decreases to zero for a sample, if it cannot be correctly classified. The $L_{EKL}$, can be defined as follows: 
\begin{equation}
\begin{multlined}
    L_{EKL} = \text{KL}[\text{Dir}(p_i|\Tilde{\alpha_i})||Dir(p_i|\textbf{1})] = \\
    log \left( \frac{\Gamma(\sum_{c=1}^{C}{\Tilde{\alpha_{ic}}})}{\Gamma(C)\prod_{c=1}^{C}{\Gamma(\Tilde{\alpha_{ic}})}} \right) +  \\ 
    \sum_{c=1}^{C}{(\Tilde{\alpha_{ic}} - 1)\Bigg[\digamma(\Tilde{\alpha_{ic}})-\digamma(\sum_{j=1}^{C}{\Tilde{\alpha_{ij}}})\Bigg]}
    \end{multlined}
\end{equation}
where $\text{KL}[.||.]$ corresponds to the Kullback-Leibler divergence, $\Tilde{\alpha_{i}}$ are the Dirichlet parameters after removal of the non-misleading evidence and $\digamma(.)$ is the digamma function.

After training the model, uncertainty and class prediction can be obtained. In evidential theory, there are several measures of uncertainty \cite{josang2018uncertainty, zhao2019quantifying}, among them are: \emph{Vacuity}, measuring the lack of evidence; and \emph{Dissonance}, measuring the conflict of evidences. These uncertainty measures are detailed as follows:
\begin{equation}
    \text{Vac}(\alpha) = \frac{C}{S}
\end{equation}
\begin{equation}
    \text{Diss}(\alpha) = \sum_{c=1}^{C}{ b_c\frac{\sum_{i \neq  c}{b_i\text{Bal}(b_i,b_c)}}{\sum_{i \neq c}{b_i}}}
\end{equation}
\begin{equation*}
\text{Bal}(b_i,b_c) =
    \begin{cases} 1 - \frac{|b_i-b_c|}{b_i+b_c} & \text{if}\; b_ib_c \neq 0 \\
                     0 &  \text{else}
    \end{cases}
\end{equation*}
where $b = \frac{e}{S}$ is the belief mass. 

\subsection{Continual Evidential Deep Learning}
The proposed Continual EDL approach (see Fig. \ref{fig:pipeline}) integrates the EDL method with the continual deep learning strategies for exemplar rehearsal and knowledge distillation to counter forgetting. Our method first applies rehearsal of data and knowledge distillation to transfer knowledge from the old model to the new one. Then motivated by Zhao et al.~\cite{zhao2020maintaining}, in a second phase, taking into account that the model still has a prediction bias towards the new classes, a Weight Aligning (WA) method is applied to correct the model trained in the first phase. 

To integrate both methods, several changes in the Continual Deep Learning architecture and loss function are necessary. Specifically, inspired by \cite{sensoy2018evidential}, the softmax activation is replaced by an Exponential (Exp) activation that appears more stable than the ReLU proposed in the original EDL implementation. Moreover, the minimization of the loss function, $L_{ECE}$ considering the evidence obtained after applying Exp activation seems similar to the typical softmax cross-entropy minimization used in classical neural networks, which ensures comparable results in terms of accuracy, but with extra information about the confidence of the prediction. As for the loss function, $L_{EKL}$ is added to preserve the model's ability to capture uncertainty. The proposed loss function is defined as follows: 
\begin{equation}
\begin{multlined}
L = \lambda_1 L_{ECE} + \lambda_2 L_{EKL} + \lambda_3 L_{KD}
\end{multlined}
\end{equation}
where $\lambda_1$, $\lambda_2$ and $\lambda_3$ weight the importance of each loss function and $L_{KD}$ is the knowledge distillation loss defined as:
\begin{equation}
\begin{multlined}
L_{KD} = \text{KL}[p^s(\tau)||p^t(\tau)] = \\
\sum_{c = 1}^{C_{old}}{p_c^t(\tau)\times(log(p_c^t(\tau))-log(p_c^s(\tau)))}
\end{multlined}
\end{equation}
with a $\tau$ parameter that acts as a temperature.

The $L_{KD}$ is incorporated to the loss function, $L$ from the second task training. It is interesting to note that the proposed loss function contemplates two regularizations, one to maintain the knowledge acquired by the old model, ($L_{KD}$) and the other - to avoid providing low uncertainty in misclassified data ($L_{EKL}$). We argue that since $L_{KD}$ allows transferring knowledge about evidence of old classes, in a continual learning approach $L_{EKL}$ should focus only on the new classes. Therefore, the $L_{EKL}$ for the t-th task is redefined as follows:
\begin{equation}
\begin{multlined}
    L_{EKL} = \text{KL}[\text{Dir}(p_{it}|\Tilde{\alpha_{it}})||Dir(p_{it}|\textbf{1})] 
    \end{multlined}
\end{equation}
where $p_{it}$ and $\alpha_{it}$ are the probability mass function and Dirichlet parameters related to the $C_{new}$ new classes for task $t$ and sample $i$, respectively. 

Finally, in the same way as \cite{he2022out}, we apply a bias correction (BC) method to reduce the output value towards biased classes. For this purpose, during inference time, the output logits are normalized based on the norm of the weight vectors in the classifier corresponding to each learned class.

\section{Validation} 
In this section, we describe the experimental setting, the evaluation procedure and metrics to validate the proposed method.
\subsection{Experimental Settings}

We follow the protocol for OOD detection evaluation used in \cite{he2022out} to compare the proposed approach with existing posthoc methods such as: MSP \cite{hendrycks2016baseline}, ODIN \cite{liang2018enhancing}, Energy \cite{liu2020energy}, Entropy \cite{kuan2022back} and MSP\_BC\_CE \cite{he2022out}. These posthoc methods are applied on top of the baseline model corresponding to the continual learning approach published in \cite{zhao2020maintaining}. For all experiments, ResNet32 \cite{he2016deep} is used as  backbone and trained for $120$ epochs with a batch size of $128$, for each incremental step. Optimization is performed using SGD with an initial learning rate of $1e-1$ that is reduced following a cosine annealing schedule, a momentum of $0.9$ and a weight decay of $5e-4$. 

The methods are evaluated on CIFAR-100 \cite{krizhevsky2009learning}, a widely used dataset in continual learning. The order of the classes is shuffled according to \cite{wiewel2022dirichlet}. Then, the dataset is divided in $5$ and $10$ tasks with $20$ and $10$ base classes and incremental step of the same size, respectively. 

As for rehearsal, we use a growing buffer that stores $20$ samples per class, herded as \cite{rebuffi2017icarl} did. Regarding the knowledge distillation, a temperature of $T = 2$ is used. With respect to the loss function, the weighting parameters were set empirically to ensure a classification performance close to the baseline. Specifically, $\lambda_1 = 0.5$ and $\lambda_2 = 0.5$ for the first task, and $\lambda_1 = 0.45$, $\lambda_2 = 0.5$ and $\lambda_3 = 0.05$ for the remaining tasks. Finally, to avoid overfitting during training, the RandAugment \cite{cubuk2020randaugment} method is used for data augmentation.

All experiments were performed using PyTorch library on a computer with a NVIDIA RTX 2080 TI graphics card. 

\subsection{Evaluation}

Detecting OOD in an unsupervised continual learning scenario involves distinguishing between unlabeled data from previously learned tasks (in-distribution or IND) and data from the new tasks (OOD) \cite{he2022out}. On the other hand, knowing whether new samples belong to old ($\text{IND}_f$), current ($\text{IND}_c$) or unseen classes (OOD) may be convenient in order to treat them differently during learning in a continuous learning framework \cite{aljundi2022continual}. Based on this, we evaluated OOD detection performance before each incremental learning step as follows:
\begin{itemize}
    \setlength{\itemsep}{3pt}
    \setlength{\parskip}{0pt}
    \setlength{\parsep}{0pt}
    \item IND vs. OOD: Ability to discriminate samples of previously learned classes with respect to samples of unseen  
    classes.
    \item $\text{IND}_c$ vs. OOD: Ability to discriminate samples of
    current classes with respect to samples of unseen classes. Useful in determining how catastrophic forgetting affects OOD detection.
    \item $\text{IND}_f$ vs. OOD: Ability to discriminate between samples of old classes with respect to unseen classes.
    \item $\text{IND}_c$ vs. $\text{IND}_f$: Ability to discriminate samples of recently learned classes with respect to old classes.
\end{itemize}

\minisection{Metrics.}
For comparative purposes, we selected the most commonly used metrics to assess both classification and OOD detection. In the case of classification, the average classification accuracy (ACA) corresponds to the accuracy of all classes seen after the $t$-th task; and the average incremental accuracy (AIA) corresponds to the average accuracy over all tasks.  In the case of OOD detection, OOD is the negative class and the rest are considered as positive classes, with the exception of the $\text{IND}_c$ vs $\text{IND}_f$ evaluation, where $\text{IND}_f$ is treated as negative class. With this in mind, in general, OOD detection performance is calculated as a binary classification problem using the area under the receiver operating characteristic curve (AUROC), the area under the accuracy-recall curve (AUPR) and the false positive rate at the 95\% of recall (FPR95). A higher value of AUROC and AUPR, and a lower value of FPR95, indicates better detection performance.

\section{Results}

In this section, the experimental results related to the detection of OOD data and the analysis of two uncertainty measures, vacuity and dissonance, are presented. 

\begin{table}
\begin{center}
\begin{tabular}{l|c|c|c|c}
\multicolumn{1}{c}{} & \multicolumn{2}{c}{Step size 10} & \multicolumn{2}{c}{Step size 20}\\
 \hline
Method & ACA & AIA & ACA & AIA\\
\hline\hline
Baseline & 50.96\% & 63.87\% & 56.36\% & 68.02\% \\
CEDL & 51.07\% & 63.16\% & 55.61\% & 66.27\% \\

\end{tabular}
\end{center}
\caption{Classification results in terms of ACA and AIA.}
\label{tab:tab1classification}
\end{table}

\begin{table*}
\begin{center}
\begin{tabular}{l|c|c|c|c|c|c}

 \multicolumn{1}{c}{} & \multicolumn{3}{c}{Step size 10} & \multicolumn{3}{c}{Step size 20}\\
 \hline
Method & AUROC $\uparrow$  & AUPR $\uparrow$ & FPR95 $\downarrow$ & AUROC $\uparrow$  & AUPR $\uparrow$ & FPR95 $\downarrow$ \\
\hline\hline
MSP \cite{hendrycks2016baseline} & 61.14\% & 86.62\% & 91.91\% & 66.34\% & 81.87\% & 90.09\% \\
ODIN \cite{liang2018enhancing} & 60.23\% & 86.25\% & 92.43\% & 65.39\% & 81.33\% & 90.73\% \\
Energy Score \cite{liu2020energy} & 60.85\% & 86.10\% & 91.29\% & 65.03\% & 80.46\% & 90.76\% \\
Entropy \cite{kuan2022back} & 61.72\% & 86.76\% & 91.05\% & 66.71\% & 82.07\% & 89.74\% \\
MSP-BC-CE \cite{he2022out} & 62.47\% & 87.11\% & 91.14\% & 67.18\% & 82.14\% & 90.00\% \\
CEDL (dissonance) & 55.74\% & 84.60\% & 95.76\% & 57.04\% & 77.40\% & 95.96\% \\
CEDL (vacuity) & \textbf{64.44\%} & \textbf{87.24\%} & \textbf{85.87\%} & \textbf{72.42\%} & \textbf{84.69\%} & \textbf{76.09\%} \\
 \hline
\end{tabular}
\end{center}
\caption{Average AUROC, AUPR and FPR95 over all tasks with step size of 10 and 20. Best results are in bold.}
\label{tab:soa}
\end{table*}

\subsection{Continual OOD detection}

In this work we perform OOD detection after each incremental step, considering two settings of 5 and 10 tasks, respectively.  Although this work focuses primarily on the OOD detection problem and not on object classification, as we can see in Table \ref{tab:tab1classification}, our approach provides comparable or slightly inferior results with respect to the baseline in terms of ACA and AIA on the test data of the classes seen. But, at the same time, it also provides the prediction's uncertainty without requiring additional computational resources.

Regarding the detection of OOD data, in our proposal we used the estimated uncertainty related to the prediction of each sample to perform this detection. Specifically through the vacuity measure, as the performance vastly outperforms the dissonance measure in detecting OOD (see Table \ref{tab:soa}). 
High vacuity means higher probability for OOD data and low vacuity is related to IND data. The obtained results are reported in terms of FPR95, AUPR and ROC for the proposed method and also for the state-of-the-art postdoc methods.  
The evaluation is performed following the protocol proposed in \cite{he2022out}, where after each task the OOD detection is performed taking into account that the test data corresponding to the learned classes are considered as IND data and the unseen classes as OOD data. The average of the results of all tasks for each metric and for two incremental step sizes is shown in Table \ref{tab:soa} for all methods. Here, we highlight the effectiveness of the proposed method, achieving a wide performance improvement in all metrics with respect to all state-of-the-art methods. Particularly, a noticeable improvement is observed for the incremental step size of 20, where a difference of 14\% of FPR95 is achieved with respect to the second-best model (MSP-BC-CE), which means that our method is able to misclassify less OOD data as IND when recall is high. In addition, with respect to the AUROC and AUPR metrics, the improvement is approximately 5.0\% and 2.5\%, respectively. 

\begin{table}
\begin{center}
\begin{tabular}{c|c|c|c}

Evaluation &Task & CEDL (vac.) & MSP-BC-CE\\
\hline\hline
&2 & 85.70\% & 93.15\% \\
$\text{IND}_f$ vs. OOD &3 & 91.10\% & 94.35\% \\
&4 & 89.75\% & 92.60\% \\
\hline\hline
&1 & 52.05\% & 85.65\% \\
$\text{IND}_c$ vs. OOD &2 & 51.85\% & 83.25\%\\
&3 & 70.15\% & 85.60\%\\
&4 & 73.15\% & 82.45\%\\
\hline\hline
&2 & 73.55\% & 88.40\% \\
$\text{IND}_c$ vs. $\text{IND}_f$ &3 & 80.20\% & 88.63\%\\
&4 & 84.45\% & 87.80\%\\
&5 & 91.39\% & 90.81\%\\

\end{tabular}
\end{center}
\caption{Performance evaluation in terms of FPR95 between different subsets of data ($\text{IND}_c$, $\text{IND}_f$ and OOD).}
\label{tab:idf_vs_ood}
\end{table}

Motivated by \cite{aljundi2022continual}, the IND data is divided into $\text{IND}_c$ and $\text{IND}_f$ to analyze the ability of the methods to distinguish between data belonging to old classes ($\text{IND}_f$), current classes ($\text{IND}_c$) and unseen classes (OOD). In Table \ref{tab:idf_vs_ood}, 
we present the results of the best and second best method in terms of FPR95 for the $\text{IND}_f$ vs. OOD, $\text{IND}_c$ vs. OOD and $\text{IND}_c$ vs. $\text{IND}_f$ evaluations.
It can be seen from the table that it is very difficult to differentiate between the data belonging to old classes with respect to data from current or unseen classes. However, in the first tasks our proposal offers a promising performance (less than 85\%) for detecting $\text{IND}_f$ data. Moreover, we can observe that the proposed method loses its good ability to distinguish between different subsets of data as the training progresses. Despite this, the proposed method is still better than MSP-BC-CE. In the case of IND vs. OOD evaluation, we observe that our proposal provides a great performance improvement with respect to MSP-BC-CE, in some tasks over 30\% difference. This performance declines when all INDs are considered, highlighting that quantified uncertainty suffers from the catastrophic forgetting of old classes.

    \begin{figure}
        \centering
        \begin{subfigure}[b]{0.235\textwidth}  
            \centering 
            \includegraphics[width=\textwidth]{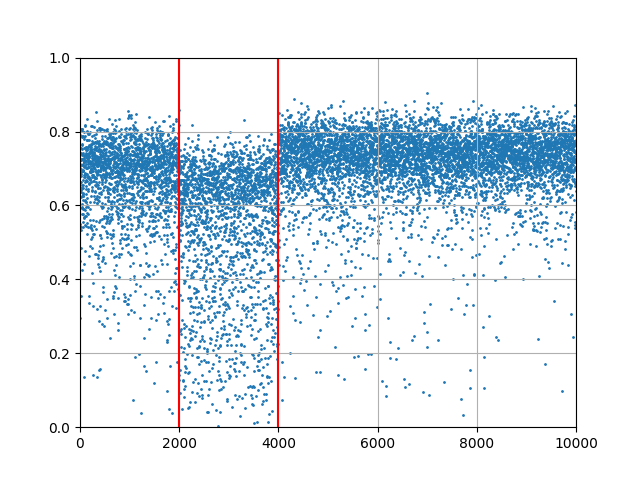}
            \caption[]%
            {{\small After Task 2}}    
        \end{subfigure}
        \hfill
        \begin{subfigure}[b]{0.235\textwidth}   
            \centering 
            \includegraphics[width=\textwidth]{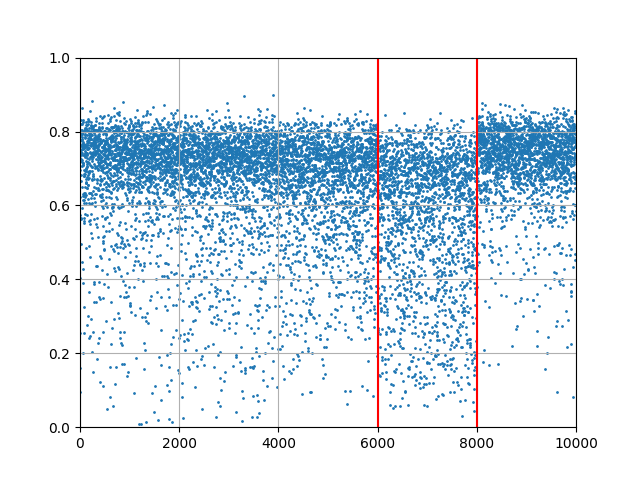}
            \caption[]%
            {{\small After Task 4}}    
        \end{subfigure} \\[-3ex]
        \vskip\baselineskip
        \begin{subfigure}[b]{0.235\textwidth}  
            \centering 
            \includegraphics[width=\textwidth]{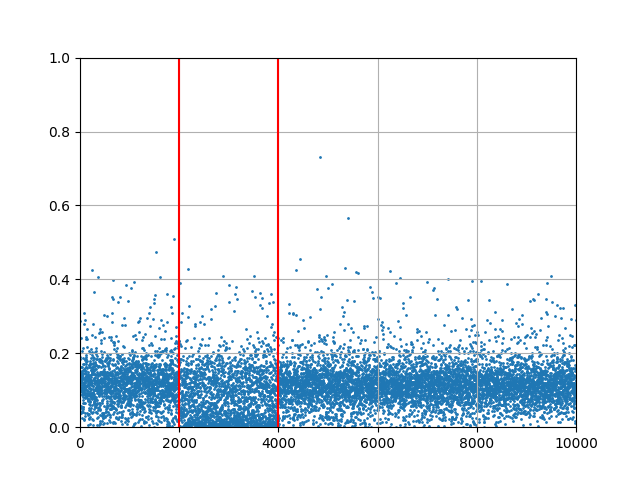}
            \caption[]%
            {{\small After Task 2}}    
        \end{subfigure}
        \hfill
        \begin{subfigure}[b]{0.235\textwidth}   
            \centering 
            \includegraphics[width=\textwidth]{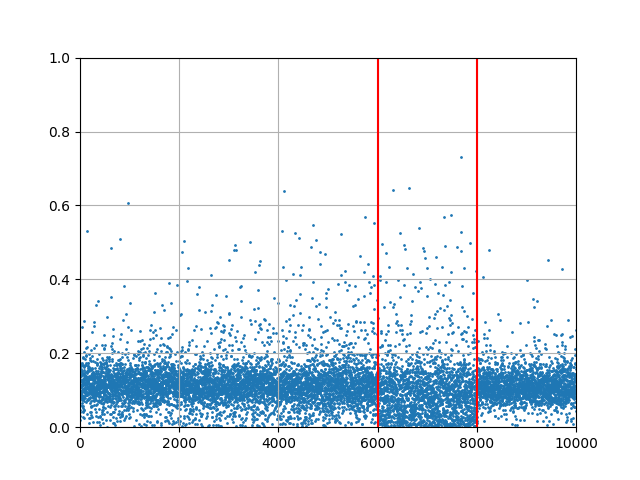}
            \caption[]%
            {{\small After Task 4}}    
        \end{subfigure}
        
        \caption[ vacuityvsdissonance ]
        {\small Vacuity (top) and dissonance (bottom) estimated for the proposed method on all test data after training of the model with task 2 and task 4. The red vertical line divides the test data in IND data (left side) with respect to the OOD data (right side). The data are grouped by old classes (left), current classes (center) and unseen classes (right).} 
        \label{fig:vac_vs_dis}
    \end{figure}

\subsection{Vacuity and dissonance analysis}

Figure \ref{fig:vac_vs_dis} shows how vacuity (top) and dissonance (bottom) change after training each task. In this experiment we use a 5-task split. Here, the resulting model after training on each task is used to quantify the uncertainty in the test data for all tasks. As expected, the uncertainty of the test data belonging to the classes of the current task is lower than that of the rest, as measured by both vacuity and dissonance. In addition, it is observed that the vacuity of the data from old classes and current classes is lower than unseen classes. However, as the training progresses, the vacuity in the data of the old classes tends to be higher and thus closer to OOD. As for dissonance, conflicting evidence is expected in the data for the old classes (mostly conflicting with the unseen classes), but in our case this only occurs in a small part of the test data. This can be seen in the model generated after task 4, where the dissonance of the old classes tends to be higher than that of the unseen ones. In most cases, the dissonance is very low. 

    \begin{figure*}
        \centering
        \begin{subfigure}[b]{0.465\textwidth}
            \centering
            \includegraphics[width=\textwidth]{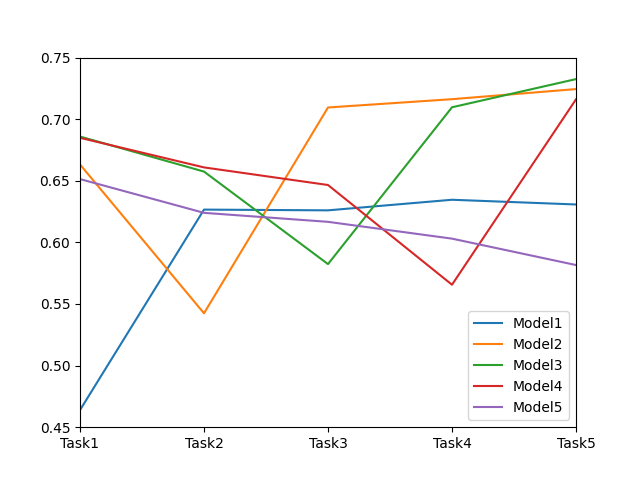}
        \end{subfigure}
        \hfill
        \begin{subfigure}[b]{0.465\textwidth}  
            \centering 
            \includegraphics[width=\textwidth]{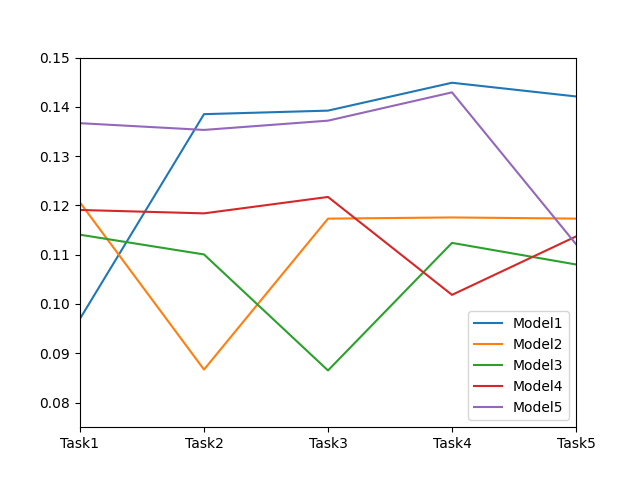}
        \end{subfigure} \\[-5ex]
        \vskip\baselineskip
        \begin{subfigure}[b]{0.465\textwidth}   
            \centering 
            \includegraphics[width=\textwidth]{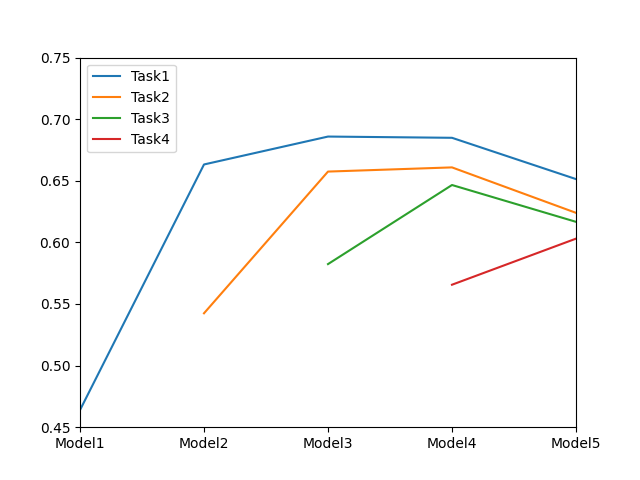}
        \end{subfigure}
        \hfill
        \begin{subfigure}[b]{0.465\textwidth}   
            \centering 
            \includegraphics[width=\textwidth]{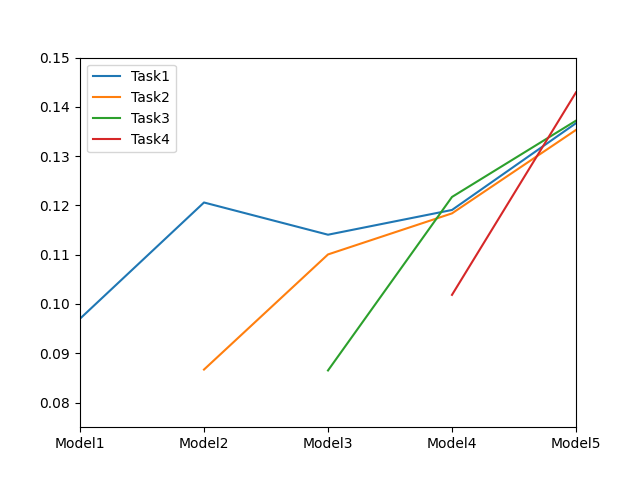}
        \end{subfigure}
        \caption[ vacuity_vs_dissonance ]
        {\small Average of Vacuity (left) and Dissonance (right) obtained by the proposed method. The i-th model corresponds to the resulting model after training with the j-th task. In the upper graphs, each line represents the results of the i-th model for the test data of the j-th task. In contrast, in the lower graphs, each line corresponds to the results of the i-th task for the j-th model.     } 
        \label{fig:vacuity_vs_dissonance_2}
    \end{figure*}

Further analysis regarding vacuity and dissonance can be performed by analyzing Figure \ref{fig:vacuity_vs_dissonance_2}. In every plot, we report the average uncertainty 
for the test data.  
In the upper graphs, each line represents the average uncertainty provided by the model trained on the i-th task for the test data for each of the tasks, separately. From that we can observe the average uncertainty with respect to the data belonging to the old (left side), current (in the task id) and unseen (right side) classes. For example, for the model trained after the second task (task id = 2), we can observe that the vacuity is low for the data belonging to the same task, high for the data belonging to unseen classes and in between for old classes. As for dissonance, for the same task, we can see that it is low for the data corresponding to the current task, high for the old classes and for the unseen classes a little lower than for the old ones. Both behaviors remain the same for all tasks and highlight the ability of the proposed model to differentiate between IND and OOD data. On the other hand, regarding the bottom graphs the line represents the uncertainty provided by the i-th model after training the i-th task for the j-th task data. In this case, we can observe how the vacuity and dissonance increase after each incremental step having the maximum difference between the current classes and the immediately unseen classes. 

    \begin{figure*}
        \centering
        \begin{subfigure}[b]{0.465\textwidth}
            \centering
            \includegraphics[width=\textwidth]{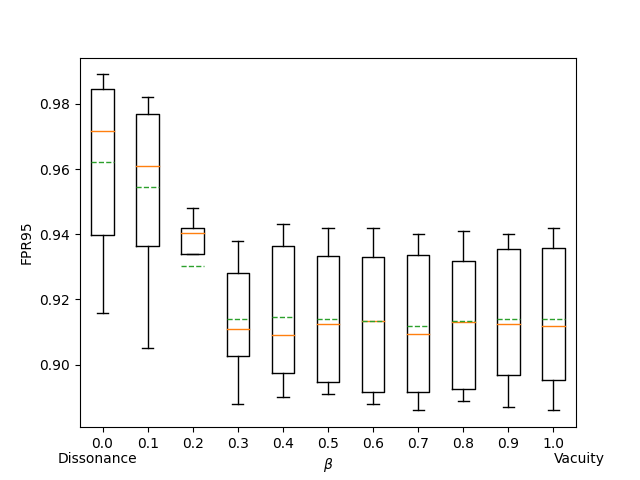}
            \caption[Network2]%
            {{\small Step size 10}}    
        \end{subfigure}
        \hfill
        \begin{subfigure}[b]{0.465\textwidth}  
            \centering 
            \includegraphics[width=\textwidth]{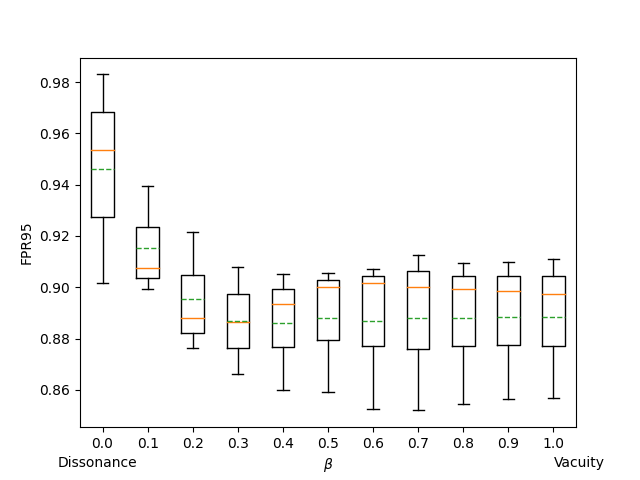}
            \caption[]%
            {{\small Step size 20}}    
        \end{subfigure}
        \caption[ vacuity_vs_dissonance ]
        {\small  Box plot illustrating the results obtained when dissonance and vacuity are combined to perform detection between $\text{IND}_f$ data and OOD data. The green and yellow lines correspond to the mean and median yield achieved in the evaluation after each increment, respectively.} 
        \label{fig:dis_vac_t}
    \end{figure*}

Considering vacuity tends to be lower for IND data than OOD and dissonance higher for IND data than OOD, we evaluate whether by combining both measures of uncertainty it would be possible to better discriminate between data belonging to old classes with respect to the unseen ones. For this, we analyze the performance in OOD detection by the proposed combination of uncertainty metrics ($CU = \beta \times \text{vacuity} + (1 - \beta) \times (1 -\text{dissonance})$) for different values of $beta$ in the range $[0-1]$ (see Figure \ref{fig:dis_vac_t}). Results are presented using incremental steps of size 10 and 20, where OOD detection is performed with respect to the test data from the previous and next tasks. The box plots show the mean and median in green and yellow lines, respectively. Here we can observe that vacuity provides much better performance than dissonance for the evaluation of $\text{IND}_f$ vs OOD. However, for some particular values of $\beta$ (e.g, $\beta = 0.3$), it is possible to improve the good performance achieved using only vacuity. We argue that this occurs, because some particular samples provide a conflicting evidence when the model is trained with new data, which can be detected with the dissonance measure and helps the model decide whether the data correspond to unseen classes or already seen by the model. 

\section{Conclusions}

In this paper, we proposed a new method, called CEDL, which integrates a deep evidential learning approach into a continual learning framework to perform OOD detection based on uncertainty estimation. For this purpose, two uncertainty measures were analyzed: vacuity and dissonance. From the experimental results, we conclude that vacuity largely outperforms the state-of-the-art for OOD detection on the CIFAR-100 dataset in both settings of 5 and 10 sequential tasks, respectively. 
In addition, it was observed that vacuity tends to be lower and dissonance higher in the old class data relative to the unseen class data. Thus, the proposed CEDL method was able to retain evidence on old class data, but evidence on current classes conflicted with the old ones. Despite the latter, both measures could be used to differentiate between old and unseen class data. 
Although vacuity alone provides a good measure for OOD data detection, we also found that the combination of vacuity and dissonance can further the OOD detection. 
Finally, it can also be observed that the ability of the model to distinguish between different subsets of data ($\text{IND}_c$, $\text{IND}_f$ and OOD) decreases after each incremental step. For this reason, in future work, we will explore improving the model's ability to cope with catastrophic forgetting in order to preserve the quality of uncertainty quantification and thus detection. 

\small{\textbf{Acknowledgements}: 
This work is funded by the Government of Chile through its ANID (No. FONDECYT INICIACIÓN 11230262), 
the Horizon EU project MUSAE (No. \museNo), 2021-SGR-01094 (AGAUR), Icrea Academia'2022 (Generalitat de Catalunya), Robo STEAM (2022-1-BG01-KA220-VET-000089434, Erasmus+ EU), DeepSense (ACE053/22/000029, ACCIÓ), DeepFoodVol (AEI-MICINN, \DFVolNo), PID2022-141566NB-I00 (AEI-MICINN), CERCA Programme / Generalitat de Catalunya, and grants PID2022-143257NB-I00, TED2021-132513B-
I00, PID2019-104174GB-I00 funded by MCIN/AEI/10.13039/501100011033, Spanish Government.}

{\small
\bibliographystyle{ieee_fullname}
\bibliography{egbib}
}

\end{document}